\documentclass[11pt]{article}

\usepackage[preprint]{acl}

\usepackage{times}
\usepackage{latexsym}

\usepackage[T1]{fontenc}

\usepackage[utf8]{inputenc}

\usepackage{microtype}

\usepackage{inconsolata}

\usepackage{graphicx}

\usepackage{amsmath}
\usepackage{amssymb}
\usepackage{booktabs}
\usepackage{multirow}
\usepackage{algorithm}
\usepackage{algpseudocode}  
\usepackage{amsthm}

\usepackage[table]{xcolor}
\usepackage{subcaption}

\title{LoaQ: Layer-wise Output Approximation Quantization}

\author{
 \textbf{Li Lin},
 \textbf{Xiaojun Wan},
\\
 Wangxuan Institute of Computer Technology, Peking University,
\\
 \small{
   efsotr\_l@stu.pku.edu.cn, wanxiaojun@pku.edu.cn
 }
}

\begin{document}
\maketitle

\begin{abstract}
A natural and intuitive idea in model quantization is to approximate each component’s quantized output to match its original. Motivated by this idea, most layer-wise post-training quantization (PTQ) methods focus on weight approximation at the linear-layer level. As a result, this local objective often yields insufficient approximations and practical deviations from the guiding intuition. Recent work has improved the approximation of linear-layer outputs within the layer-wise PTQ framework, but such refinements remain inadequate for achieving alignment with the full-model output.
Based on a deeper understanding of the structure of mainstream LLMs, we propose \textbf{LoaQ}, which incorporates output-matching factors when quantizing linear layers within the layer-wise PTQ framework. It better aligns with this intuition and can feature a simple closed-form solution, making it orthogonal to existing techniques and readily integrable into existing quantization pipelines.
Experiments on the LLaMA and Qwen model families demonstrate that LoaQ performs effectively in both weight-only and weight-activation quantization. By integrating seamlessly with existing quantization strategies, it further enhances overall quantization quality and shows strong potential to advance the frontier of post-training quantization.\footnote{The code will be available for download}
\end{abstract}

\section{Introduction} \label{sec:intro}

In recent years, large-scale models have achieved remarkable breakthroughs in performance, yet these advances come with significant memory overhead, posing challenges for practical deployment. Model compression techniques~\cite{model_compress} have emerged as a promising solution to this problem, with post-training quantization (PTQ)~\cite{ptq} being one of the most impactful approaches. Within PTQ methods, layer-wise PTQ has attracted particular attention. By formulating quantization at the linear-layer level, layer-wise PTQ simplifies both the quantization process and the optimization objectives. Despite this simplification, layer-wise PTQ consistently delivers strong empirical performance. Moreover, this approach offers significant efficiency advantages: it can quantize large-scale models within a short time using only a single GPU. Beyond its standalone effectiveness, layer-wise PTQ also serves as an excellent initialization for end-to-end PTQ frameworks, accelerating their convergence and thereby further enhancing the performance. These properties collectively make layer-wise PTQ an important contributor to advancing the performance boundaries of quantized models.

Although layer-wise PTQ methods, from classic GPTQ~\cite{gptq} to more recent MagR~\cite{magr}, have achieved notable success, they still primarily focus on weight approximation. As a result, to preserve the layer-wise setting, the only additional information they leverage is the input to each linear layer. Consequently, they are confined to activation-aware weight approximation, which diverges from the natural objective of output approximation. Recent approaches, including GPTAQ~\cite{gptaq}, Qronos~\cite{qronos}, and QEP~\cite{qep}, overcome this limitation through an error-compensation strategy. Specifically, they modify weights to compensate for the mismatch between pre- and post-quantization inputs. By explicitly incorporating this compensation, these methods enable a more output-oriented optimization at the linear-layer level and ultimately achieve more accurate linear-layer output approximation.

While previous methods approximate outputs at the linear-layer level, this scope remains confined and fails to account for residual connections and RMSNorm, making it insufficient for model-level output approximation. To address these issues, we extend the perspective to the sub-block level, allowing us to naturally account for these components. Building on this idea, we introduce LoaQ, a layer-wise PTQ method that explicitly targets sub-block output approximation, thereby overcoming the limitations of linear-layer methods and aligning more closely with model-level output approximation.
Extensive experiments on LLaMA~\cite{llama2,llama3} and Qwen~\cite{qwen3} model families confirm that LoaQ consistently preserves higher quantization accuracy across diverse tasks and quantization settings. Furthermore, LoaQ integrates seamlessly with other quantization strategies, including the Hadamard transform and NeUQI~\cite{neuqi}, and delivers additional improvements.

Our main contributions are as follows:
\begin{itemize}
    \item We present a perspective based on output approximation and analyze the characteristics of the resulting errors, thereby offering potential directions for designing more effective quantization strategies.
    \item We propose LoaQ, a novel approach that effectively addresses the output approximation problem within the layer-wise PTQ framework.
    \item We validate the strong performance and potential of LoaQ through extensive experiments on diverse tasks and quantization settings, advancing the performance limits of post-training quantization, particularly for layer-wise methods.
\end{itemize}

\section{Related Works} \label{sec:related_work}

\paragraph{Layer-wise PTQ} 
Layer-wise PTQ independently optimizes each linear layer to preserve overall model performance, enabling simpler optimization procedures and more targeted algorithm design. 
Building on the layer-wise paradigm, prior work often improves PTQ by incorporating complementary techniques that reduce quantization error or stabilize optimization within each layer. Representative approaches include SmoothQuant~\citep{smoothquant} and AWQ~\citep{awq}, which balance the numerical scales of weights and activations to alleviate quantization error; the Hadamard transform adopted in QuaRot~\citep{quarot} and QuIP\#~\citep{quip} to smooth the weight distribution; NeUQI~\citep{neuqi}, which improves the initialization of uniform quantization parameters; and BCD~\citep{cdquant}, which applies coordinate descent to refine the quantized weights of each linear layer.

\paragraph{Error-compensated PTQ}
Methods such as GPTAQ~\cite{gptaq}, Qronos~\cite{qronos}, and QEP~\cite{qep} address the fact that existing layer-wise PTQ methods fail to model the discrepancy between pre-quantization inputs and post-quantization inputs when quantizing a linear layer, resulting in accumulated error. However, these error compensation methods still do not consider the presence of residual connections, and thus, error accumulation along residual connection paths remains unresolved.

\section{Preliminaries} \label{sec:preliminary}

\subsection{Linear-layer Loss} \label{sec:linear-layer_loss}

The linear-layer loss is designed to guide the optimization toward reducing the quantization error in the linear layer output, formulated as
\begin{equation} \label{eq:linear-layer_loss}
\mathcal{L}(Q)=\| X(Q - W)\|_F^2,
\end{equation}
where $W$ denotes the weight of the linear layer, $Q$ denotes its quantized counterpart, and $X$ denotes the input, also referred to as the activation, to the linear layer. 
Evidently, the linear-layer loss is independent across the columns of $W$, and thus can be decomposed as follows:
\begin{equation} \label{eq:linear-layer_loss_per-column}
\mathcal{L}(q)=\|X(q-w)\|_2^2,
\end{equation}
where $w$ is a single column of $W$, and $q$ denotes its quantized counterpart. 
Moreover, $H=X^\top X$ is the Hessian matrix of the per-column linear-layer loss $\mathcal{L}(q)$.

\begin{figure*}[t]
    \centering
    \includegraphics[width=0.8\linewidth]{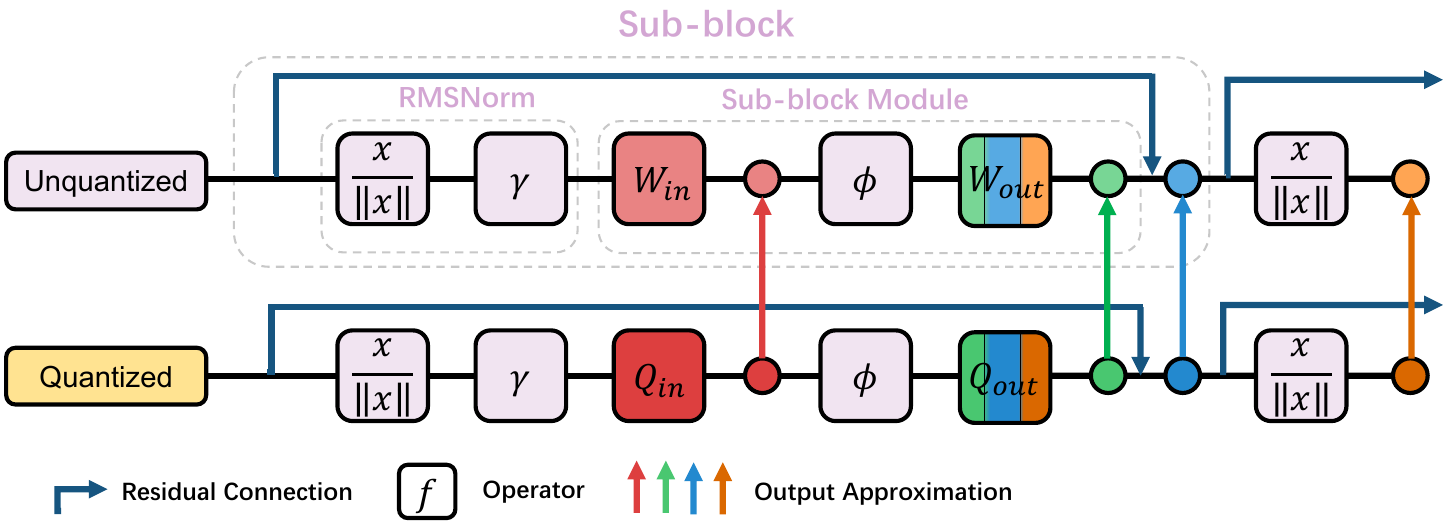}
    \caption{The overall diagram of \textbf{LoaQ}. Each colored vertical bidirectional arrow denotes the corresponding approximate output obtained when quantizing the weight matrix of the same color.}
    \label{fig:loaq}
\end{figure*}

\subsection{GPTQ} \label{sec:ldlq_algo}

To minimize the loss function in Eq.~\ref{eq:linear-layer_loss}, the core idea of the GPTQ algorithm is to quantize the weight matrix sequentially in a row-wise manner, compensating for the quantization errors of previously processed rows using the remaining unquantized rows. The formulation is given as follows:
\begin{align} \label{eq:ldlq_idea}
\widetilde{W} &= W + T (Q - \widetilde W), \nonumber \\
Q &= \mathcal{Q}(\widetilde W)
\end{align}
where $\mathcal{Q}$ denotes the quantization mapping, and $T$ denotes the compensation matrix, which is a strictly lower-triangular matrix that ensures each row accounts only for the quantization errors of the preceding rows.
As stated in~\citet{quip}, the optimal compensation matrix $T$ is $D^{-1}(L - D)$, where $L$ and $D$ are derived from the Cholesky decomposition of $H = L L^\top$ and $D$ is the diagonal matrix formed by the diagonal elements of $L$.

\subsection{GPTAQ}

Building on GPTQ, the GPTAQ algorithm~\citep{gptaq} extends this approach by introducing an compensation mechanism to better align quantized layer outputs with their original counterparts. The loss function for each linear layer is formulated as  
\begin{equation} \label{eq:gptaq_linear_layer_loss}
\mathcal{L}(Q)=\|XQ-X'W \|_F^2,
\end{equation}
where $X'$ denotes the input to the original linear layer. 
The formulation of GPTAQ is 
\begin{align} \label{eq:gptaq_formulation}
\widetilde{W} &= W + D^{-1}(L - D)(Q - \widetilde{W}) \nonumber \\
&\quad\quad + L(M_L \odot (L^\top (M_L \odot  C)))\widetilde{W}, \nonumber \\
Q &= \mathcal{Q}(\widetilde{W}),&
\end{align}
where $C = X^\top(X' - X)$ serves as a correction term that compensates for the mismatch between the original and quantized layer inputs, $M_L$ denotes a strictly lower-triangular masking matrix with ones below the diagonal, and $\odot$ denotes element-wise multiplication. We can readily observe that the GPTAQ algorithm completely ignores the information in the upper-triangular part of $C$.

In addition, directly applying the formulation in Eq.~\ref{eq:gptaq_formulation} yields worse performance than GPTQ. Therefore, \citet{gptaq} introduce a tuning parameter $\alpha$, as shown in their GitHub repository.
\begin{align} \label{eq:gptaq_formulation_w_alpha}
\widetilde{W} &= W + D^{-1}(L - D)(Q - \widetilde{W}) \nonumber \\
&\quad\quad + \alpha L(M_L \odot (L^\top (M_L \odot  C)))\widetilde{W}, \nonumber \\
Q &= \mathcal{Q}(\widetilde{W}).
\end{align}

\subsection{Sub-block} \label{sec:sub-block}

In mainstream LLM architectures, each Transformer block consists of a self-attention sub-block and an MLP sub-block. As shown in Figure~\ref{fig:loaq}, each sub-block typically includes an RMSNorm, a sub-block module, and a residual connection adding the input to the output. The sub-block can be uniformly represented in a simplified form:
\begin{align} \label{eq:simplified-sub-block}
X_k^{\mathrm{out}} &= \phi\big(\mathrm{RMSNorm}(h_k)W_{\mathrm{in}}\big), \\
h_{k+1} &= h_k + X_k^{\mathrm{out}} W_{\mathrm{out}}.
\end{align}

For the self-attention module, $W_{\mathrm{in}} = [W_q; W_k; W_v]$ are the input projections to queries, keys, and values. The nonlinearity $\phi$ denotes the masked attention mechanism, which typically includes rotary positional encoding (RoPE), causal attention, and may also incorporate QK-Norm. The output projection is given by $W_{\mathrm{out}} = W_o$.

For the MLP module, $W_{\mathrm{in}} = [W_{\mathrm{gate}}; W_{\mathrm{up}}]$ corresponds to the projections used in the gated feedforward path. The nonlinearity $\phi$ denotes an element-wise activation that gates the $W_{\mathrm{up}}$ projection using $W_{\mathrm{gate}}$ in the expanded hidden space. The projection back to the model dimension is given by $W_{\mathrm{out}} = W_{\mathrm{down}}$.

\section{Methodology} \label{sec:method}

In this section, we introduce \textbf{LoaQ}, short for \textbf{L}ayer-wise \textbf{O}utput \textbf{A}pproximation \textbf{Q}uantization. 
LoaQ builds on the layer-wise post-training quantization framework, which typically quantizes only $W_{\text{in}}$ and $W_{\text{out}}$, and extends it by incorporating additional factors during quantization to achieve more accurate approximations. 
As shown in Figure~\ref{fig:loaq}, LoaQ is organized into three components with hierarchical approximation objectives: linear-layer, sub-block, and sub-block followed by RMSNorm. The design transitions from linear-layer-level approximation to block-wise approximation, thereby imposing increasingly stringent constraints on aligning the quantized model with the original model.

\subsection{Linear Layer Output Approximation} \label{sec:linear-layer-OA}

We aim to quantize the two linear layers $W_{\text{in}}$ and $W_{\text{out}}$ such that, as indicated by the red and green unidirectional arrows in Figure~\ref{fig:loaq}, the quantized outputs approximate their original counterparts as closely as possible. For each linear layer, we optimize the loss function in Eq.~\ref{eq:gptaq_linear_layer_loss}.
By treating $X$ analogously to a common factor, we obtain an equivalent loss function for minimization:
\begin{equation} \label{obj:linear-layer-OA-solution}
\mathcal{L}(Q)= \|X(Q-X^\dagger X'W)) \|_F^2,
\end{equation}
where $X^\dagger = (X^\top X)^{-1}X^\top$ is left pseudoinverse. This loss function can be further simplified as 
\begin{equation} \label{obj:linear-layer-OA-solution-simple}
\mathcal{L}(Q) = \|X \big(Q - (I + H^{-1}C)W\big)\|_F^2.
\end{equation}
Therefore, updating $W$ to $(I + H^{-1}C)W$ converts the problem into the loss function in Eq.~\ref{eq:linear-layer_loss}, which can be directly solved using the GPTQ algorithm.

\subsection{Sub-block Output Approximation} \label{sec:sub-block-OA}

Ideally, methods that approximate only the linear-layer output can at best match the sub-block output. However, in the mainstream LLM architecture, the resulting errors accumulate through the residual connection. Accordingly, as indicated by the blue unidirectional arrow in Figure~\ref{fig:loaq}, LoaQ accounts for this effect when quantizing $W_\mathrm{out}$ by matching the sub-block output rather than only the linear-layer output.
Given that $h$ and $h'$ denote the hidden states of the quantized sub-block and the original sub-block, respectively, we obtain the loss function
\begin{align}\label{obj:sub-block-OA}
\mathcal{L}(Q)= \|(h + X Q) - (h' + X' W)\|_F^2.
\end{align}
By applying the same procedure used to derive Eq.~\ref{obj:linear-layer-OA-solution-simple}, we obtain the following equivalent form:
\begin{align}\label{obj:sub-block-OA-solution}
\arg \min_Q \|X\big(Q - \big(&(I + H^{-1} C) W + \nonumber \\ &H^{-1} X^\top (h' - h)\big) \big)\|_F^2.
\end{align}
Updating $W$ to $(I + H^{-1} C) W + H^{-1} X^\top (h'-h)$ converts the problem into the loss function in Eq.~\ref{eq:linear-layer_loss}.

\begin{figure}[t]
    \centering
    \includegraphics[width=\linewidth]{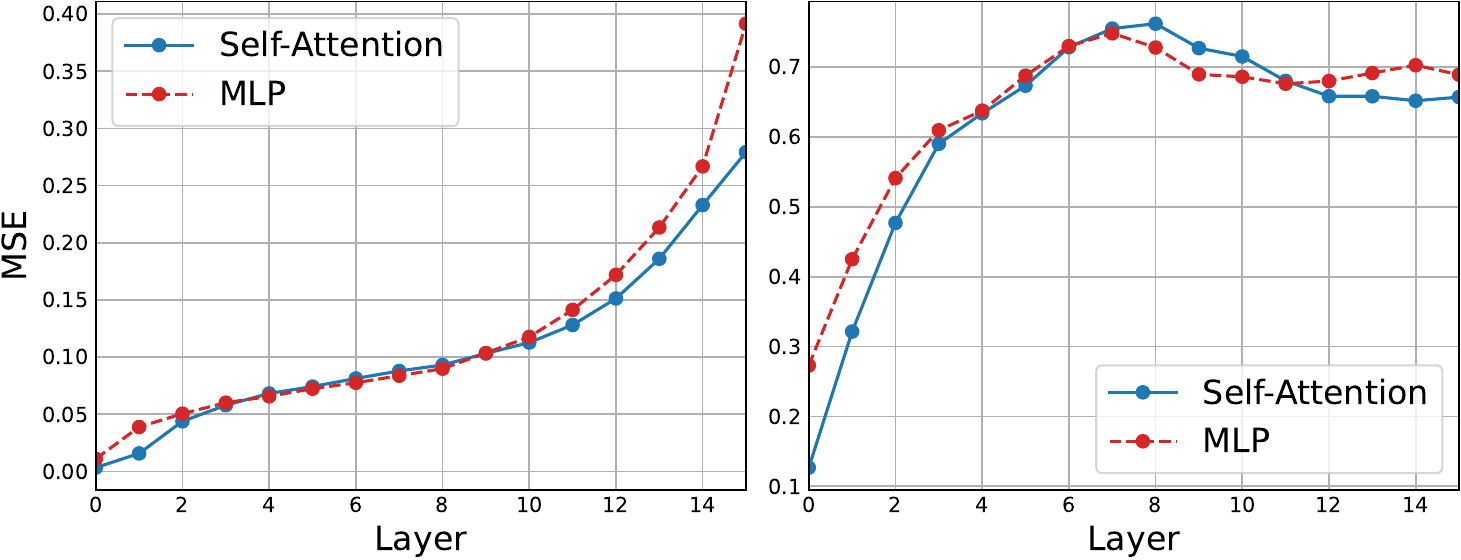}
    \caption{Results under 3-bit quantization on LLaMA 3.2 1B, showing the mean squared error (MSE) of sub-blocks' outputs. The left plot reports the MSE of unnormalized outputs, while the right plot reports the MSE of normalized outputs.}
    \label{fig:pre-error}
\end{figure}

\subsection{Normalized Sub-block Output Approximation} \label{sec:norm-OA}

Based on the above understanding of the mainstream LLM architecture, the presence of residual connections leads to the accumulation of quantization error as the depth increases, as shown in the left plot of Figure~\ref{fig:pre-error}. This raises the question of how the model remains stable despite the accumulation of quantization error. Upon closer examination, we observed that an RMSNorm layer is consistently applied before the input to each sub-block module and the final language model head. As illustrated in the right plot of Figure~\ref{fig:pre-error}, this normalization mechanism prevents the mean squared error (MSE) between normalized outputs from growing beyond a certain depth.

Since a normalization layer follows each sub-block, it is the normalized output, rather than the raw sub-block output, that directly influences subsequent linear layers. Consequently, as indicated by the orange unidirectional arrow in Figure~\ref{fig:loaq}, aligning the normalized outputs of the quantized and original models becomes essential. To this end, we refine the objective to explicitly target the normalized sub-block outputs:
\begin{align} \label{obj:normalized-sub-block-OA}
\mathcal{L}(Q)= \| \rho(h + X Q) - \rho(h' + X' W) \|_F^2, 
\end{align}
where $\rho(\cdot)$ is coloum-wise root mean square norm operator.
To simplify the expression, we rewrite the normalization term as $\rho(h + X Q) = R(h + X Q)(h + X Q)$, where $R(\cdot)$ denotes the column-wise root mean square scaling operator. To decouple the quantized terms, we replace $Q$ with the original weight $W$ inside the scaling function $R(\cdot)$, and can then be transformed into the form:
\begin{align} \label{obj:simplified-normalized-sub-block-OA}
& \mathcal{L}(Q)= \| \big( R(h + X W) h + (R(h + X W)X) Q \big) \nonumber \\
&  - \big( R(h' + X' W)h + (R(h' + X' W)X') W \big) \|_F^2
\end{align}
This yields the same form as Eq.~\ref{obj:sub-block-OA}, allowing us to apply the same solution.

\subsection{Tuning Parameters}

\begin{figure}[t]
    \centering
    \includegraphics[width=0.9\linewidth]{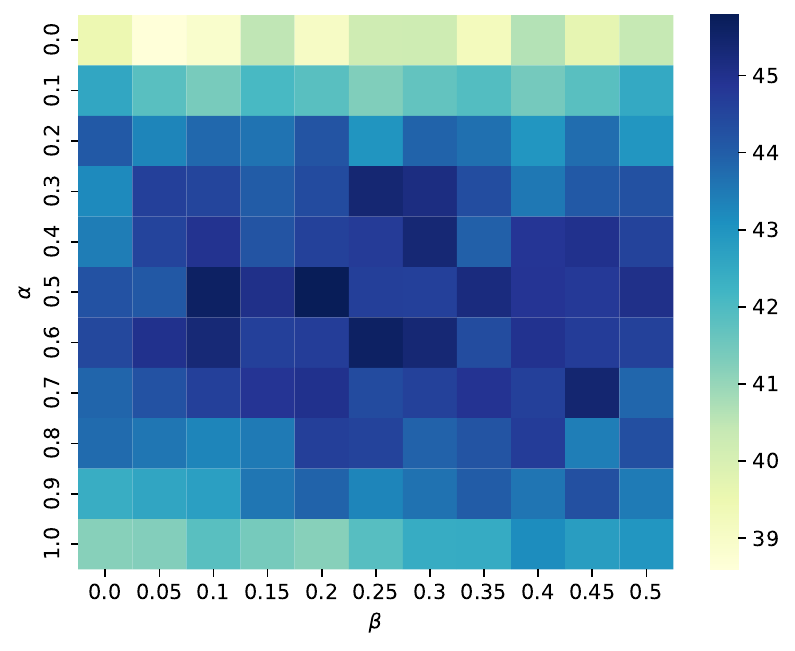}
    \caption{Results under 3-bit quantization on LLaMA 3.2 1B, reporting average accuracy across different values of $\alpha$ and $\beta$.}
    \label{fig:avg_acc_heatmap}
\end{figure}

For a single linear layer, the range of adjustable corrections is inherently limited. When the required correction is excessively large, it cannot be accommodated regardless of the adjustment strategy. In practice, the differences between hidden states are often large. When $h' - h$ becomes too large, it is difficult to find a suitable $Q$ such that $h + XQ$ closely approximates $h' + X'W$. To address this issue, we introduce tuning parameters $\alpha$ and $\beta$ and change the update rule of $W$ to  $(I + \alpha H^{-1} C) W + \beta H^{-1} X^\top (h' - h)$. The detailed LoaQ pipeline is shown in Algorithm~\ref{alg:loaq_pipeline}.
As shown in Figure~\ref{fig:avg_acc_heatmap}, the optimal performance is achieved when $\alpha$ takes intermediate values such as 0.4, 0.5, or 0.6, rather than 1. 

\begin{algorithm}[ht]
\caption{LoaQ Pipeline}
\label{alg:loaq_pipeline}
\begin{algorithmic}[1]
\Require Weights $\{W_\mathrm{in}^{(\ell)},W_\mathrm{out}^{(\ell)}\}_{\ell=1}^L$, calibration inputs, bit‐width $k$, adjustment parameters $\alpha,\beta$
\Ensure Quantized weights $\{Q^{(\ell)}\}_{\ell=1}^L$
\State Run forward to collect embedding outputs $h, h'$
\For{$\ell=1$ \textbf{to} $L$}
  \For{$W$ \textbf{in} $\{W_{\mathrm{in}}^{(\ell)},W_{\mathrm{out}}^{(\ell)}\}$}
    \State Run forward to collect $X,X'$
    \If{$W=W_{\mathrm{in}}^{(\ell)}$} \Comment{quantize $W_\mathrm{in}$}
      \State $H\leftarrow X^\top X$
      \State $C\leftarrow X^\top(X'-X)$ \Comment{Correction term from Section~\ref{sec:linear-layer-OA}}
      \State $\Delta=\textbf{0}$
    \Else \Comment{quantize $W_\mathrm{out}$}
      \State $s'\leftarrow R(h'+X'W)$
      \State $s\leftarrow R(h+XW)$ \Comment{Rescale for each token from Section~\ref{sec:norm-OA}}
      \State $H\leftarrow (sX)^\top (sX)$
      \State $C\leftarrow (sX)^\top(s'X'-sX)$ 
      \State $\Delta\leftarrow H^{-1} (sX)^\top\bigl(s' h'-sh\bigr)$ \Comment{Correction term from Section~\ref{sec:sub-block-OA}}
    \EndIf
    \State $\widetilde W\leftarrow\bigl(I+\alpha H^{-1}C\bigr)W+\beta\Delta$
    \State Quantize: $Q^{(\ell)}\leftarrow\mathrm{GPTQ}(\widetilde W;H;k)$
  \EndFor
  \State Run forward to get updated sub-block outputs $h, h'$
\EndFor
\State \Return $\{Q^{(\ell)}\}_{\ell=1}^L$
\end{algorithmic}
\end{algorithm}

\begin{table*}[t]
\renewcommand{\arraystretch}{0.95}
\small
\centering
\begin{tabular}{ccc|cc|cccccc}
\toprule
\textbf{Model} & \textbf{Size} & \textbf{Method} & \textbf{Wiki2↓} & \textbf{C4↓} & \textbf{ArcC↑} & \textbf{ArcE↑} & \textbf{HellaS↑} & \textbf{PiQA↑} & \textbf{WinoG↑} & \textbf{Acc↑} \\

\midrule
\multirow[c]{12}{*}{LLaMA 2} & \multirow[c]{4}{*}{7B} & GPTQ & 6875 & 2237 & \textbf{22.27} & 25.25 & 26.02 & 53.10 & 50.91 & 35.51 \\
 &  & GPTAQ & 1269 & 246 & 21.76 & 26.89 & 25.74 & 53.32 & 47.12 & 34.97 \\
 &  & Qronos & 1505 & 306 & 20.65 & 27.82 & 25.96 & 53.70 & 49.09 & 35.44 \\
 &  & LoaQ & \textbf{214} & \textbf{67.31} & 19.45 & \textbf{30.56} & \textbf{28.71} & \textbf{55.98} & \textbf{54.46} & \textbf{37.83} \\
\cmidrule(lr){2-11}
 & \multirow[c]{4}{*}{13B} & GPTQ & 4235 & 580 & \textbf{20.90} & 25.84 & 25.72 & 50.33 & 50.91 & 34.74 \\
 &  & GPTAQ & 145 & 62.05 & 20.56 & 26.89 & 26.99 & 53.43 & \textbf{51.85} & 35.95 \\
 &  & Qronos & 352 & 110 & 20.82 & 26.68 & 26.78 & 52.18 & 50.12 & 35.31 \\
 &  & LoaQ & \textbf{111} & \textbf{47.26} & 20.14 & \textbf{33.54} & \textbf{31.61} & \textbf{59.52} & 50.75 & \textbf{39.11} \\
\cmidrule(lr){2-11}
 & \multirow[c]{4}{*}{70B} & GPTQ & 71.05 & 40.17 & \textbf{21.59} & 28.03 & 26.96 & 53.48 & 53.12 & 36.63 \\
 &  & GPTAQ & \textbf{40.09} & 28.37 & \textbf{21.59} & 28.91 & 28.98 & 54.08 & 51.46 & 37.00 \\
 &  & Qronos & 66.52 & 41.75 & 19.45 & 34.85 & 29.78 & 60.23 & 53.04 & 39.47 \\
 &  & LoaQ & 41.62 & \textbf{28.03} & 19.97 & \textbf{35.98} & \textbf{32.53} & \textbf{60.88} & \textbf{54.85} & \textbf{40.84} \\
\midrule
\multirow[c]{8}{*}{LLaMA 3} & \multirow[c]{4}{*}{8B} & GPTQ & $>$1e4 & $>$1e4 & 22.01 & 25.04 & 25.87 & \textbf{54.62} & 48.15 & 35.14 \\
 &  & GPTAQ & $>$1e4 & 4409 & 20.99 & \textbf{25.29} & 25.88 & 52.99 & 48.46 & 34.72 \\
 &  & Qronos & $>$1e4 & $>$1e4 & \textbf{22.95} & 25.21 & 25.59 & 54.35 & 48.62 & 35.35 \\
 &  & LoaQ & \textbf{5497} & \textbf{528} & 22.53 & 24.41 & \textbf{26.37} & 53.05 & \textbf{51.54} & \textbf{35.58} \\
\cmidrule(lr){2-11}
 & \multirow[c]{4}{*}{70B} & GPTQ & $>$1e4 & $>$1e4 & 21.25 & 25.17 & 25.72 & 52.34 & 51.14 & 35.12 \\
 &  & GPTAQ & $>$1e4 & $>$1e4 & 22.70 & 24.87 & \textbf{25.78} & 53.21 & 51.22 & 35.56 \\
 &  & Qronos & \textbf{12147} & \textbf{9187} & 20.22 & 25.17 & 25.73 & 53.10 & 49.80 & 34.81 \\
 &  & LoaQ & $>$1e4 & $>$1e4 & \textbf{23.21} & \textbf{26.39} & 25.75 & \textbf{53.32} & \textbf{52.41} & \textbf{36.22} \\
\midrule
\multirow[c]{12}{*}{Qwen 3} & \multirow[c]{4}{*}{8B} & GPTQ & 7151 & 2969 & 20.82 & 25.42 & 25.25 & 50.54 & 51.22 & 34.65 \\
 &  & GPTAQ & 500 & 171 & \textbf{21.16} & 25.29 & 25.14 & 50.82 & 47.04 & 33.89 \\
 &  & Qronos & 263 & 129 & 19.97 & 27.90 & 27.01 & 54.68 & 47.83 & 35.48 \\
 &  & LoaQ & \textbf{171} & \textbf{82.90} & 18.52 & \textbf{33.75} & \textbf{29.96} & \textbf{58.65} & \textbf{52.33} & \textbf{38.64} \\
\cmidrule(lr){2-11}
 & \multirow[c]{4}{*}{14B} & GPTQ & 1916 & 888 & \textbf{23.46} & 24.79 & 25.07 & 51.36 & 47.83 & 34.50 \\
 &  & GPTAQ & 224 & 101 & 21.67 & 25.13 & 25.81 & 48.69 & 48.93 & 34.05 \\
 &  & Qronos & 141 & 76.67 & 19.28 & 32.32 & 28.52 & 57.40 & 51.07 & 37.72 \\
 &  & LoaQ & \textbf{85.40} & \textbf{54.57} & 21.08 & \textbf{36.41} & \textbf{32.47} & \textbf{60.72} & \textbf{53.35} & \textbf{40.81} \\
\cmidrule(lr){2-11}
 & \multirow[c]{4}{*}{32B} & GPTQ & 8444 & 2843 & 21.67 & 24.75 & 25.61 & 51.90 & 49.57 & 34.70 \\
 &  & GPTAQ & 321 & 117 & \textbf{23.55} & 24.07 & 26.62 & 50.38 & 51.14 & 35.15 \\
 &  & Qronos & 156 & 77.23 & 23.46 & 28.11 & 29.63 & 54.13 & 48.86 & 36.84 \\
 &  & LoaQ & \textbf{107} & \textbf{60.44} & 19.88 & \textbf{34.43} & \textbf{33.33} & \textbf{60.45} & \textbf{52.25} & \textbf{40.07} \\
\bottomrule
\end{tabular}
\caption{Perplexity on Wiki2 and C4, along with the average zero-shot accuracy across five benchmarks (reported as \textbf{Acc}), for various quantized models using 2-bit channel-wise quantization. The arrows \textbf{↑}/\textbf{↓} indicate whether higher or lower is better.}
\label{tab:main-2bit}
\end{table*}

\section{Experiment} \label{sec:experiment}

\subsection{Baselines and Evaluation} \label{sec:baseline}

We experiment with our proposed LoaQ on three commonly-used LLM families, covering different sizes: LLaMA 2 (7B, 13B, 70B)~\cite{llama2}, LLaMA 3 (8B, 70B)~\cite{llama3} and Qwen 3 (8B, 14B, 32B)~\cite{qwen3}.
Under the weight-only quantization setting, we compare LoaQ with the baseline GPTQ~\cite{gptq} and two error compensation methods, Qronos~\cite{qronos} and GPTAQ~\cite{gptaq}. We also evaluate scenarios that incorporate effective quantization techniques such as NeUQI~\cite{neuqi} and the Hadamard transform to assess the effectiveness of LoaQ under these enhancements. In addition, LoaQ is assessed under the weight-activation quantization setting. A brief introduction to these techniques is provided in Appendix~\ref{appendix:quant_tech}.
Following the previous work, all the quantized models are evaluated by measuring perplexity on the WikiText2 (\textbf{Wiki2})~\cite{wikitext2} and \textbf{C4}~\cite{c4} validation sets, and zero-shot accuracy on five benchmarks: ARC-easy (\textbf{ArcE}), ARC-challenge (\textbf{ArcC})~\cite{arc}, \textbf{PiQA}~\cite{piqa}, HellaSwag (\textbf{HellaS})~\cite{hellaswag} , and WinoGrande (\textbf{WinoG})~\cite{winogrande}.


\subsection{Implementation Detail} \label{sec:implement}

For weight-only quantization, we evaluate the 2-bit channel-wise quantization and the 3-bit channel-wise quantization.
For weight-activation quantization, we adopt the W4A4 setting, where weights are quantized to 4 bits channel-wise and activations to 4 bits token-wise dynamically, with quantization parameters determined by Min-Max~\cite{integer-only}.
During calibration, following prior work, we draw 128 samples of 2048 tokens each from the C4 dataset for the LLaMA 2, LLaMA 3, and Qwen 3 families, with a fixed random seed to ensure reproducibility. We use the LM Evaluation Harness~\citep{eval-harness} to evaluate zero-shot accuracy on five benchmarks.
All experiments are conducted on NVIDIA A40.
Our proposed LoaQ reformulates the problem as Eq.~\ref{sec:linear-layer_loss}. We use GPTQ in our experiments, but the formulation also supports other optimization methods.
More implementation details are provided in Appendix~\ref{appendix:implementation_details}.

\subsection{Main Result} \label{sec:main-result}

Under the highly challenging 2-bit channel-wise quantization scenario, LoaQ generally surpasses the baseline GPTQ and two strong counterparts, GPTAQ and Qronos, as shown in Table~\ref{tab:main-2bit}. For LLaMA 3 70B, all methods perform poorly due to its inherent difficulty in 2-bit quantization. In contrast, LoaQ demonstrates strong and consistent performance across different model families and sizes.

As the bit-width increases, the performance gap between methods narrows, as shown in Table~\ref{tab:main-3bit} in Appendix. LoaQ continues to show clear advantages under the 3-bit setting, demonstrating robustness across model architectures, while in some cases performing on par with others and still offering slight benefits in specific scenarios.

\subsection{Discussion}

\subsubsection{Hadmard Transform} \label{sec:hadamard-transform}

As shown in Table~\ref{tab:had}, we further evaluate LoaQ combined with the Hadamard transform, which smooths weight and activation distributions to mitigate quantization error. Compared with methods that focus only on linear-layer output approximation such as GPTAQ and Qronos, LoaQ substantially delivers better performance across all models. In the challenging 2-bit setting it remains stable where other methods fail, and even on Qwen 3 8B under the 3-bit setting, where perplexities are relatively higher, it still achieves the best overall accuracy. These results highlight the robustness and wide applicability of LoaQ.

\begin{table}[t]
\small
\centering
\begin{tabular}{c@{\hskip 4.55pt}|@{\hskip 4.55pt}c@{\hskip 4.55pt}c@{\hskip 4.55pt}c|c@{\hskip 4.55pt}c@{\hskip 4.55pt}c}
\toprule
\textbf{Method} & \textbf{Wiki2↓} & \textbf{C4↓} & \textbf{Acc↑} & \textbf{Wiki2↓} & \textbf{C4↓} & \textbf{Acc↑} \\
\midrule
& \multicolumn{3}{c|}{2-bit} & \multicolumn{3}{c}{3-bit} \\
\midrule
\multicolumn{7}{c}{LLaMA 2 7B} \\
\midrule
GPTQ & 759 & 277 & 34.95 & 7.36 & 9.04 & 59.68 \\
GPTAQ & 64.88 & 39.36 & 37.14 & 6.93 & 8.54 & 60.35 \\
Qronos & 77.19 & 48.52 & 37.05 & 6.88 & 8.46 & 60.22 \\
LoaQ & \textbf{44.65} & \textbf{23.76} & \textbf{41.81} & \textbf{6.73} & \textbf{8.31} & \textbf{60.91} \\
\midrule
\multicolumn{7}{c}{LLaMA 3 8B} \\
\midrule
GPTQ & 7883 & 1809 & 35.34 & 12.62 & 14.36 & 57.44 \\
GPTAQ & 175 & 94.41 & 35.20 & 9.57 & 12.85 & 60.24 \\
Qronos & 129 & 85.50 & 35.11 & 9.71 & 12.77 & 60.56 \\
LoaQ & \textbf{83.79} & \textbf{49.43} & \textbf{38.35} & \textbf{9.30} & \textbf{12.56} & \textbf{61.52} \\
\midrule
\multicolumn{7}{c}{Qwen 3 8B} \\
\midrule
GPTQ & 2592 & 997 & 35.05 & 12.32 & 15.53 & 59.42 \\
GPTAQ & 103 & 59.24 & 35.22 & 12.13 & 15.35 & 59.59 \\
Qronos & 106 & 51.08 & 36.08 & \textbf{11.89} & \textbf{15.19} & 60.18 \\
LoaQ & \textbf{45.45} & \textbf{33.53} & \textbf{41.26} & 12.05 & 15.27 & \textbf{62.25} \\
\bottomrule
\end{tabular}
\caption{Results on LLaMA 2 7B, LLaMA 3 8B, and Qwen 3 8B for 2-bit and 3-bit channel-wise quantization with the Hadamard transform.}
\label{tab:had}
\end{table}

\subsubsection{Parameter Initialization} \label{sec:neuqi}
    
We further evaluated LoaQ combined with NeUQI, an effective quantization parameter initialization method, and assessed its performance with and without the equally effective Hadamard transform, as shown in Table~\ref{tab:neuqi}. Under the 2-bit channel-wise setting, LoaQ consistently outperforms GPTQ, GPTAQ, and Qronos, while under the 3-bit channel-wise setting, it achieves stable improvements. The results obtained through these combinations with effective quantization techniques highlight LoaQ's strong potential to serve as an important component of an ultimate PTQ solution, particularly for layer-wise PTQ under low-bit-width configurations.

\begin{table}[t]
\centering
\small
\begin{tabular}{c@{\hskip 4.3pt}|c@{\hskip 4.3pt}c@{\hskip 4.3pt}c|c@{\hskip 4.3pt}c@{\hskip 4.3pt}c}
\toprule
\textbf{Method} & \textbf{Wiki2↓} & \textbf{C4↓} & \textbf{Acc↑} & \textbf{Wiki2↓} & \textbf{C4↓} & \textbf{Acc↑} \\
\midrule
& \multicolumn{3}{c|}{No Transform} & \multicolumn{3}{c}{Hadamard Trasform} \\
\midrule
\multicolumn{7}{c}{LLaMA 2 7B} \\
\midrule
GPTQ & 17.44 & 17.62 & 47.88 & 12.41 & 13.22 & 52.91 \\
GPTAQ & 13.91 & 14.33 & 49.36 & 9.63 & 10.75 & 54.79 \\
Qronos & 12.97 & 13.04 & 52.36 & 9.99 & 10.90 & 55.08 \\
LoaQ & \textbf{11.10} & \textbf{11.47} & \textbf{53.21} & \textbf{8.62} & \textbf{9.82} & \textbf{56.44} \\
\midrule
\multicolumn{7}{c}{LLaMA 3 8B} \\
\midrule
GPTQ & 46.03 & 33.21 & 48.74 & 18.15 & 22.03 & 51.93 \\
GPTAQ & 43.34 & 27.84 & 50.69 & 16.21 & 18.48 & 53.88 \\
Qronos & 42.38 & 27.65 & 50.86 & 15.73 & 17.57 & 53.78 \\
LoaQ & \textbf{28.88} & \textbf{22.87} & \textbf{52.24} & \textbf{12.43} & \textbf{15.34} & \textbf{55.51} \\
\midrule
\multicolumn{7}{c}{Qwen 3 8B} \\
\midrule
GPTQ & 64.02 & 36.12 & 42.55 & 17.15 & 19.74 & 59.64 \\
GPTAQ & 205 & 103 & 38.34 & 170 & 98.48 & 37.81 \\
Qronos & 50.54 & 31.44 & 48.27 & 16.06 & 18.94 & 58.10 \\
LoaQ & \textbf{28.74} & \textbf{23.70} & \textbf{51.40} & \textbf{15.59} & \textbf{18.52} & \textbf{60.01} \\
\bottomrule
\end{tabular}
\caption{Results on LLaMA 2 7B, LLaMA 3 8B, and Qwen 3 8B for 2-bit channel-wise quantization with and without the Hadamard transform, combined with NeUQI.}
\label{tab:neuqi}
\end{table}
\begin{table}[t]
\centering
\small
\begin{tabular}{c@{\hskip 4.3pt}|c@{\hskip 4.3pt}c@{\hskip 4.3pt}c|c@{\hskip 4.3pt}c@{\hskip 4.3pt}c}
\toprule
\textbf{Method} & \textbf{Wiki2↓} & \textbf{C4↓} & \textbf{Acc↑} & \textbf{Wiki2↓} & \textbf{C4↓} & \textbf{Acc↑} \\
\midrule
& \multicolumn{3}{c|}{No Transform} & \multicolumn{3}{c}{Hadamard Trasform} \\
\midrule
\multicolumn{7}{c}{LLaMA 2 7B} \\
\midrule
GPTQ & 429 & 445 & 36.98 & 6.15 & 7.76 & 62.38 \\
GPTAQ & 902 & 617 & 35.65 & 6.20 & 7.73 & 61.93 \\
Qronos & 579 & 159 & 35.22 & 6.38 & 7.91 & 61.64 \\
LoaQ & \textbf{62.98} & \textbf{31.06} & \textbf{39.33} & \textbf{6.11} & \textbf{7.65} & \textbf{62.47} \\
\midrule
\multicolumn{7}{c}{LLaMA 3 8B} \\
\midrule
GPTQ & 323 & 422 & 38.33 & 8.06 & 11.38 & 62.53 \\
GPTAQ & 52.08 & 50.59 & 44.16 & \textbf{7.85} & \textbf{11.16} & 63.38 \\
Qronos & 411 & 166 & 36.50 & 8.68 & 11.77 & 61.96 \\
LoaQ & \textbf{28.38} & \textbf{30.76} & \textbf{45.09} & 7.91 & 11.18 & \textbf{63.99} \\
\midrule
\multicolumn{7}{c}{Qwen 3 8B} \\
\midrule
GPTQ & 2919 & 2241 & 34.72 & \textbf{11.02} & \textbf{14.63} & 63.57 \\
GPTAQ & 3380 & 2043 & 35.15 & 11.07 & 14.69 & 64.53 \\
Qronos & 1943 & 1158 & 35.12 & 12.16 & 15.73 & 64.78 \\
LoaQ & \textbf{1187} & \textbf{643} & \textbf{35.38} & 11.33 & 15.00 & \textbf{65.31} \\
\bottomrule
\end{tabular}
\caption{Results of W4A4 with and without Hadamard Transform on LLaMA 2 7B, LLaMA 3 8B, and Qwen 3 8B. }
\label{tab:w4a4}
\end{table}

\subsubsection{Weight-Activation Quantization} \label{sec:wa-quant}

For the W4A4 setting, as shown in Table~\ref{tab:w4a4}, LoaQ remains highly competitive. Without the Hadamard transform, LoaQ demonstrates clear performance gains over a wide range of models. When the Hadamard transform is applied, GPTAQ and Qronos experience performance drops on some models, whereas LoaQ still delivers slight improvements. These results demonstrate that LoaQ is effective in both weight-only and weight-activation quantization, confirming its broad applicability to model quantization.

\subsubsection{Ablation Study} \label{sec:ablation}

\begin{figure}[t]
    \centering
    \includegraphics[width=\linewidth]{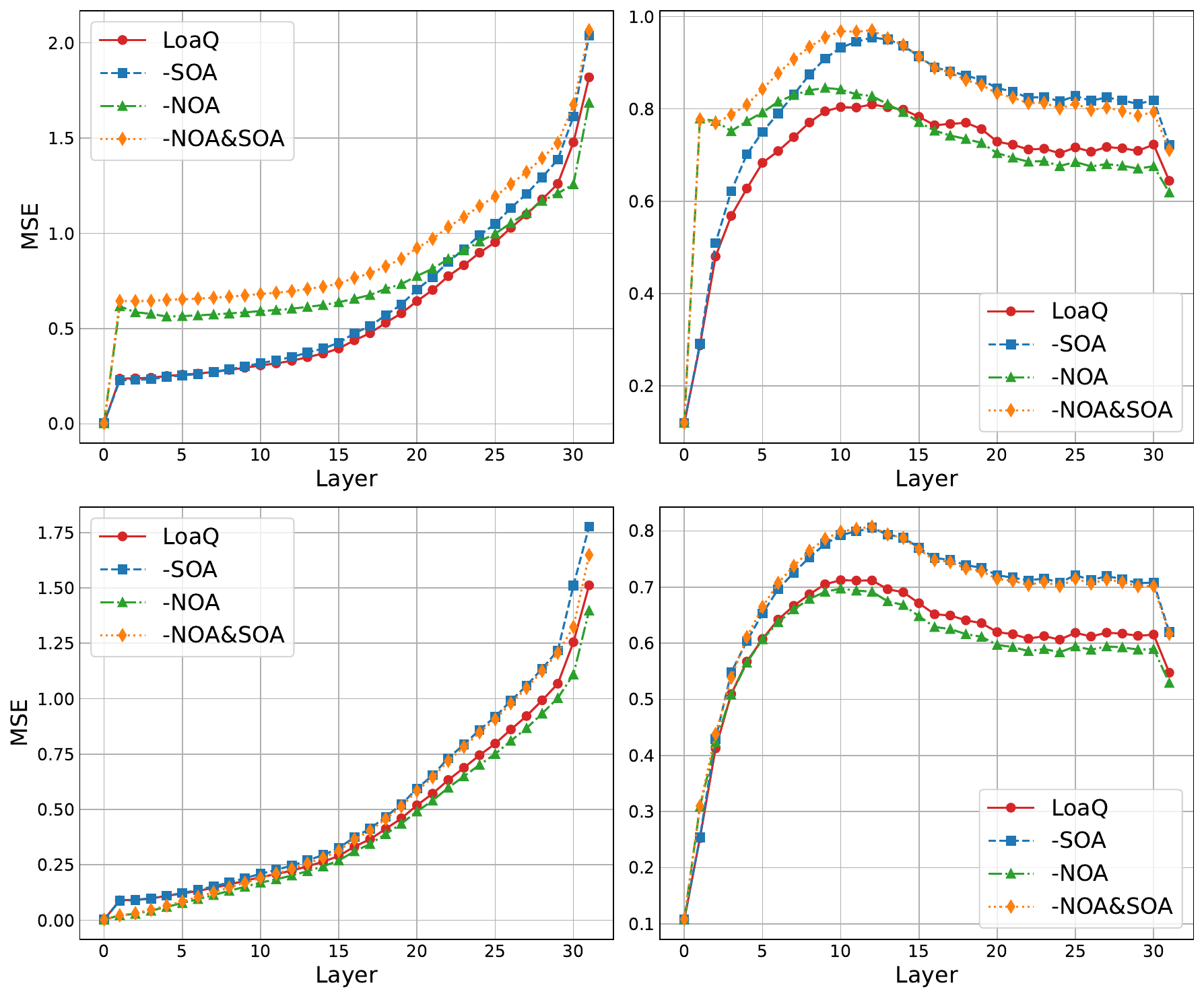}
    \caption{Results of 2-bit channel-wise quantization on LLaMA 2 7B, presenting the mean squared error (MSE) of Transformer block outputs excluding outlier tokens. The top two plots show the MSE without the Hadamard transform, while the bottom two plots show the MSE with the Hadamard transform. For each row, the left plot corresponds to unnormalized outputs, and the right plot corresponds to normalized outputs.}
    \label{fig:ablation}
\end{figure}

We further evaluate the individual contributions of the Sub-block Output Approximation (SOA) and the Normalized Sub-block Output Approximation (NOA) in LoaQ through an ablation study on LLaMA 2 7B under 2-bit channel-wise quantization. We examine three configurations: removing SOA, removing NOA, and removing both SOA and NOA. The results in Table~\ref{tab:ablation} demonstrate that both SOA and NOA provide clear benefits. Moreover, as shown in Figure~\ref{fig:ablation}, the top two plots without the Hadamard transform reveal that SOA markedly reduces the mean squared error of hidden states across all layers, while NOA substantially lowers errors in the early layers. The corrective effect of NOA appears to diminish with increasing layer depth, likely because accumulated quantization errors weaken its impact. In contrast, the bottom two plots with the Hadamard transform show a considerable overall improvement. Under this setting, the influence of NOA is less directly reflected in the error, while SOA still exhibits a clear reduction in errors. These results confirm that both SOA and NOA are effective, and their combination, especially when paired with the Hadamard transform, further enhances quantization performance.

\begin{table}[t]
\small
\begin{tabular}{c@{\hskip 4.3pt}|c@{\hskip 4.3pt}c@{\hskip 4.3pt}c|c@{\hskip 4.3pt}c@{\hskip 4.3pt}c}
\toprule
\textbf{Method} & \textbf{Wiki2↓} & \textbf{C4↓} & \textbf{Acc↑} & \textbf{Wiki2↓} & \textbf{C4↓} & \textbf{Acc↑} \\
\midrule
& \multicolumn{3}{c|}{No Transform} & \multicolumn{3}{c}{Hadamard Trasform} \\
\midrule
LoaQ & \textbf{214} & \textbf{67.31} & \textbf{37.83} & 44.65 & 23.76 & \textbf{41.81} \\
-SOA & 225 & 69.93 & 37.57 & 45.79 & 26.81 & 40.69 \\
-NOA & 1757 & 250 & 35.04 & \textbf{36.50} & \textbf{23.12} & 40.83 \\
\quad-SOA & 328 & 95.15 & 35.85 & 44.00 & 28.32 & 40.06 \\
GPTQ & 6875 & 2237 & 35.51 & 759 & 277 & 34.95 \\
\bottomrule
\end{tabular}
\caption{Ablation study of the SOA and NOA with and without the Hadamard Transform on LLaMA 2 7B using 2-bit channel-wise quantization.}
\label{tab:ablation}
\end{table}

\subsubsection{Quantization Overhead} \label{sec:overhead}

The overhead, reported in minutes in Table~\ref{tab:overhead}, represents the time required to execute the quantization algorithms for 3-bit quantization of the Qwen 3 model family on an NVIDIA A40 GPU. LoaQ, along with GPTAQ and Qronos, needs to store and process both pre- and post-quantization inputs as well as auxiliary information, resulting in overheads of the same order of magnitude. Compared with GPTAQ and Qronos, LoaQ incurs a slightly higher cost because it leverages richer contextual information and performs additional computations at multiple stages of the quantization process. This modest increase in quantization time is well justified by the substantial performance improvements achieved by LoaQ, reflecting the effectiveness of its design.

\begin{table}[t]
  \centering
  \small
  \begin{tabular}{cccc}
    \toprule
    \multirow{2}{*}{\textbf{Method}} & \multicolumn{3}{c}{\textbf{Size}} \\
    \cmidrule(lr){2-4}
     & \textbf{8B} & \textbf{14B} & \textbf{32B} \\
    \midrule
    GPTQ   &  11.48 &  20.63 &  55.23 \\
    GPTAQ  &  17.92 &  34.13 & 100.00 \\
    Qronos &  17.45 &  33.27 &  97.87 \\
    LoaQ   &  19.68 &  37.78 & 108.03 \\
    \bottomrule
  \end{tabular}
  \caption{Overhead in minutes of 3-bit channel-wise quantization on the Qwen 3 family using an NVIDIA A40 GPU.}
  \label{tab:overhead}
\end{table}

\section{Conclusion} \label{sec:conclusion}

Across various models, settings, and combinations with advanced quantization techniques, LoaQ consistently outperforms methods that rely solely on linear-layer output approximation, such as GPTAQ and Qronos. This demonstrates the value of designing layer-wise PTQ based on the structure of LLMs, leading to quantization methods that better align with optimization objectives. These results also underscore LoaQ's potential to enhance the effectiveness of layer-wise PTQ. In future work, we plan to further explore techniques such as error compensation and output approximation, as well as investigate phenomena related to quantization errors, to develop more effective strategies for layer-wise PTQ.

\section*{Limitations}

In this work, we focus on corrections within the layer-wise PTQ framework, particularly GPTQ. Accordingly, the evaluation is restricted to the layer-wise PTQ framework, and the resulting conclusions may not generalize to methods outside the layer-wise PTQ setting. Moreover, the LoaQ method alone still exhibits a performance gap compared to full-precision (FP16/BF16) models and therefore requires other complementary techniques for support. In addition, the performance of quantized models can vary significantly depending on the distribution of the training data, and their behavior in more specific application scenarios remains unclear.

\section*{Ethical considerations}

This work does not raise any ethical concerns. All open-source large language models and datasets used in our work are publicly available. All resources are used in compliance with their respective licenses and solely for research purposes. We carefully review the datasets used in our work to ensure that they do not contain personally identifiable information or offensive content. For the use of AI assistants, we employ them to polish descriptive content and generate plotting code.


\bibliography{custom}

\appendix

\section{Quantization Techniques} \label{appendix:quant_tech}

In this section, we briefly introduce the quantization techniques involved in the Experiment section.

\subsection{Qronos}

The Qronos algorithm~\cite{qronos} optimizes the same loss function as GPTAQ, as defined in Eq.~\ref{eq:gptaq_linear_layer_loss}.
The formulation of Qronos is 
\begin{align} \label{eq:qronos_formulation}
\widetilde{W} &= W + D^{-1}(L - D)(Q - \widetilde{W}) \nonumber \\
    &\quad\quad +I_{:,2:} (H_{2:, 2:})^{-1}C_{2:, :}W, \nonumber \\
Q &= \mathcal{Q}(\widetilde{W}+e^{(1)}(H_{1,1}^{-1}C_{1,:}W)),
\end{align}
where $I$ denotes the identity matrix and $e^{(1)}$ denotes the first standard basis vector.  Consequently, the term $e^{(1)}H_{1,1}^{-1}C_{1,:}W$ applies a correction to the first row of $\widetilde{W}$ by adding $H_{1,1}^{-1}C_{1,:}W$.
In addition, directly applying the formulation in Eq.~\ref{eq:qronos_formulation} yields worse performance than GPTQ. Therefore, \citet{qronos} set $X' = X$ at the beginning of each block.

\subsection{Hadamard Transform}

The Hadamard transform\footnote{See \url{https://en.wikipedia.org/wiki/Hadamard_matrix}}, which is used in methods such as QuaRot and QuIP\#, smooths the distributions of activations and weights, enabling more effective quantization while maintaining high efficiency due to the special structure of the Hadamard matrix, which allows fast computation. Specifically, it employs an orthogonal matrix $R$ with entries $\pm 1$, enabling the transformations $X' = XR$ and $W' = R^{-1}W$. Here, $X'$ and $W'$ serve as the transformed activations and weights that are subsequently used for quantization. When the transform is applied to a vector of length $n$, the computational complexity is $O(n \log n)$ on a single worker and can be reduced to $O(\log n)$ with sufficient workers.

\subsection{NeUQI}

NeUQI is a quantization parameter initialization method designed for uniform quantization. It overcomes the limitations of the conventional Min-Max framework, particularly the restriction that the zero-point must be an integer, resulting in a highly efficient initialization method with significant performance improvements.

\section{Implementation Details} \label{appendix:implementation_details}

\subsection{GPTQ}

Our proposed LoaQ reformulates the problem as Eq.~\ref{eq:linear-layer_loss}, which we solve using the LDLQ algorithm~\citep{quip}, which is mathematically equivalent to GPTQ. To improve numerical stability, we add a small diagonal damping term equal to one percent of the average value of the diagonal entries of $H$ to the Hessian matrix $H$ to ensure that it is invertible.
To further improve performance in the 2-bit setting, following \citet{gptq}, we execute the LDLQ algorithm by permuting the row order of the weight matrix in descending order of the diagonal entries of $H$, rather than using the original row order.

\subsection{Hyperparameters} \label{appendix:hyperparameter}

We perform a hyperparameter grid search by first tuning $\alpha$ and then tuning $\beta$.  
Specifically, $\alpha$ ranges from 0 to 1 in increments of 0.1 and $\beta$ ranges from 0 to 0.5 in increments of 0.05.  
\begin{align}
\alpha &\in \{\,0.1 \times i \mid i = 0,1,2,\ldots,10\,\}, \\
\beta  &\in \{\,0.05 \times j \mid j = 0,1,2,\ldots,10\,\}.
\end{align}
$\alpha = 0.4$ and $\beta = 0.15$ are empirically selected as effective values based on a search conducted on the LLaMA and Qwen families.

\section{Supplementary Results} \label{appendix:supplementary_results}

This appendix presents supplementary quantization results. 
Detailed results on 3-bit channel-wise quantization for LLaMA and Qwen families are shown in Table~\ref{tab:main-3bit}.

\begin{table*}[hbtp]
\renewcommand{\arraystretch}{0.95}
\small
\centering
\begin{tabular}{ccc|cc|cccccc}
\toprule
\textbf{Model} & \textbf{Size} & \textbf{Method} & \textbf{Wiki2↓} & \textbf{C4↓} & \textbf{ArcC↑} & \textbf{ArcE↑} & \textbf{HellaS↑} & \textbf{PiQA↑} & \textbf{WinoG↑} & \textbf{Acc↑} \\
\midrule
\multirow[c]{12}{*}{LLaMA 2} & \multirow[c]{4}{*}{7B} & GPTQ & 8.39 & 9.98 & 33.02 & 65.15 & 48.98 & 73.50 & 63.77 & 56.89 \\
 &  & GPTAQ & 7.61 & 9.15 & 34.81 & 68.52 & 50.06 & 74.32 & \textbf{65.98} & 58.74 \\
 &  & Qronos & 7.80 & 9.17 & 33.45 & 67.26 & 50.19 & 74.05 & 64.25 & 57.84 \\
 &  & LoaQ & \textbf{7.46} & \textbf{8.92} & \textbf{39.51} & \textbf{71.13} & \textbf{51.57} & \textbf{75.41} & 65.27 & \textbf{60.58} \\
\cmidrule(lr){2-11}
 & \multirow[c]{4}{*}{13B} & GPTQ & 6.45 & 8.00 & \textbf{41.81} & 73.82 & 55.08 & 76.44 & 68.27 & 63.08 \\
 &  & GPTAQ & 6.31 & 7.84 & 40.96 & \textbf{75.29} & 55.30 & 76.66 & \textbf{69.93} & \textbf{63.63} \\
 &  & Qronos & 6.39 & 7.88 & 41.04 & 74.96 & 54.50 & 76.12 & 69.69 & 63.26 \\
 &  & LoaQ & \textbf{6.26} & \textbf{7.76} & 41.21 & 74.96 & \textbf{55.47} & \textbf{77.48} & 68.90 & 63.60 \\
\cmidrule(lr){2-11}
 & \multirow[c]{4}{*}{70B} & GPTQ & 4.84 & 6.58 & 49.40 & \textbf{80.72} & \textbf{60.80} & 79.54 & 74.43 & 68.98 \\
 &  & GPTAQ & 4.81 & 6.58 & 47.70 & 79.42 & 58.23 & 79.76 & \textbf{74.59} & 67.94 \\
 &  & Qronos & 4.80 & 6.52 & 49.83 & 79.84 & 60.79 & 79.98 & 74.51 & \textbf{68.99} \\
 &  & LoaQ & \textbf{4.77} & \textbf{6.51} & \textbf{50.17} & 79.38 & 59.95 & \textbf{80.63} & 74.43 & 68.91 \\
\midrule
\multirow[c]{8}{*}{LLaMA 3} & \multirow[c]{4}{*}{8B} & GPTQ & 24.52 & 22.75 & 27.22 & 51.14 & 44.27 & 64.74 & 59.43 & 49.36 \\
 &  & GPTAQ & 17.79 & 18.05 & 32.59 & 62.04 & 46.69 & 69.59 & 64.88 & 55.16 \\
 &  & Qronos & 33.17 & 23.76 & 28.50 & 56.02 & 47.88 & 69.64 & 63.69 & 53.15 \\
 &  & LoaQ & \textbf{12.53} & \textbf{15.01} & \textbf{32.85} & \textbf{63.38} & \textbf{50.23} & \textbf{73.45} & \textbf{66.38} & \textbf{57.26} \\
\cmidrule(lr){2-11}
 & \multirow[c]{4}{*}{70B} & GPTQ & \textbf{1909} & \textbf{931} & 20.90 & 25.21 & \textbf{26.15} & 52.50 & 51.38 & 35.23 \\
 &  & GPTAQ & $>$1e4 & 8500 & 20.39 & \textbf{25.67} & 25.70 & \textbf{53.10} & 50.51 & 35.08 \\
 &  & Qronos & $>$1e4 & 7331 & 20.90 & 25.08 & 25.85 & 52.61 & 51.07 & 35.10 \\
 &  & LoaQ & $>$1e4 & $>$1e4 & \textbf{22.35} & 24.75 & 25.82 & 52.99 & \textbf{51.78} & \textbf{35.54} \\
\midrule
\multirow[c]{12}{*}{Qwen 3} & \multirow[c]{4}{*}{8B} & GPTQ & 14.74 & 17.36 & 32.59 & 52.69 & 47.97 & 71.55 & 56.51 & 52.26 \\
 &  & GPTAQ & 14.03 & 16.77 & 37.37 & 66.84 & 48.45 & 72.63 & 61.64 & 57.39 \\
 &  & Qronos & 14.21 & 16.69 & 38.74 & 65.91 & 48.62 & 71.55 & 62.83 & 57.53 \\
 &  & LoaQ & \textbf{13.72} & \textbf{16.68} & \textbf{41.55} & \textbf{74.24} & \textbf{49.50} & \textbf{74.32} & \textbf{64.88} & \textbf{60.90} \\
\cmidrule(lr){2-11}
 & \multirow[c]{4}{*}{14B} & GPTQ & 12.43 & 14.88 & 43.52 & 71.46 & 53.19 & 75.14 & 63.85 & 61.43 \\
 &  & GPTAQ & \textbf{11.78} & 14.73 & 45.90 & 74.28 & 53.29 & 76.06 & 68.43 & 63.59 \\
 &  & Qronos & 12.06 & 14.75 & \textbf{47.18} & \textbf{76.89} & 53.74 & \textbf{76.66} & 67.40 & \textbf{64.38} \\
 &  & LoaQ & 11.86 & \textbf{14.70} & 46.25 & 75.72 & \textbf{54.47} & 75.79 & \textbf{68.90} & 64.22 \\
\cmidrule(lr){2-11}
 & \multirow[c]{4}{*}{32B} & GPTQ & 11.71 & 14.06 & 40.87 & 67.09 & 55.95 & 74.92 & 62.59 & 60.28 \\
 &  & GPTAQ & 10.87 & 13.44 & \textbf{48.29} & 74.45 & 56.87 & 75.95 & 67.80 & 64.67 \\
 &  & Qronos & 11.32 & 13.52 & 44.45 & 73.19 & 56.76 & \textbf{76.99} & 68.43 & 63.96 \\
 &  & LoaQ & \textbf{10.56} & \textbf{13.10} & 45.56 & \textbf{75.25} & \textbf{57.58} & 76.88 & \textbf{69.69} & \textbf{64.99} \\
\bottomrule
\end{tabular}
\caption{Results of 3-bit channel-wise quantization, applied to the LLaMA 2, LLaMA 3, and Qwen 3 model families.}
\label{tab:main-3bit}
\end{table*}

\end{document}